\documentclass[letterpaper]{article}
\usepackage{aaai19}
\usepackage{times}
\usepackage{helvet}
\usepackage{graphicx}
\usepackage{amssymb}
\usepackage{courier}
\begin{document}
%
\title{Situational Grounding within Multimodal Simulations}
\author{James Pustejovsky and Nikhil Krishnaswamy\\
Department of Computer Science\\
Brandeis University\\
{\tt \{jamesp,nkrishna\}@brandeis.edu}\\
}
\maketitle
\begin{abstract}
In this paper, we argue that simulation platforms enable a novel type of embodied spatial reasoning, one facilitated by a formal model of object and event semantics that renders the continuous quantitative search space of an open-world, real-time environment tractable.  We provide examples for how a semantically-informed AI system can exploit the precise, numerical information provided by a game engine to perform qualitative reasoning about objects and events, facilitate learning novel concepts from data, and communicate with a human to improve its models and demonstrate its understanding.  We argue that simulation environments, and game engines in particular, bring together many different notions of ``simulation" and many different technologies to provide a highly-effective platform for developing both AI systems and tools to experiment in both machine and human intelligence.
\end{abstract}

\section{Introduction}


The concept of simulation has played an important role in both  AI and cognitive science for over 40 years. There are two distinct uses for the term {\it simulation}, particularly as used in computer science and AI. First, simulation can be used as a description for {\it testing a computational model}. That is, variables in a model are set and the model is run, such that the consequences of all possible computable configurations become known. Examples of such simulations include models of  climate change, the tensile strength of materials,  models of biological pathways, and so on. We refer to this as {\it computational simulation modeling}, where the goal is to arrive at the best model by using simulation techniques. 

Simulation can also refer to an environment which allows a user to interact with objects in a ``virtual or simulated world", where the agent is embodied as a dynamic point-of-view or avatar in a proxy situation.  Such simulations are used for training humans in scripted scenarios, such as flight simulators, battle training, and of course, in video gaming: in these contexts,  the software and gaming world assume an embodiment of the agent in the environment, either as a first-person restricted POV (such as a first-person shooter or RPG), or an omniscient movable embodied perspective (e.g., real-time or turn-based strategy). We refer to such approaches as {\it situated embodied simulations}.  The goal is to simulate an agent within a situation. 

Simulation has yet another meaning, however. 
Starting with Craik \shortcite{craik1943nature}, we encounter the notion that agents carry a  mental model of external reality in their heads. Johnson-Laird \shortcite{johnson1987could} develops his own theory of a mental model, which represents a situational possibility, capturing what is
common to all the different ways in which the situation may
occur \cite{johnson2002conditionals}. This is used to drive
inference and reasoning, both factual and counterfactual. Simulation Theory, as developed in philosophy of mind, has focused on the role that ``mind reading'' plays in modeling the mental representations of other agents and the content of their communicative acts \cite{gordon1986folk,goldman1989interpretation,heal1996simulation,goldman2006simulating}. Simulation semantics (as adopted within cognitive linguistics and  practiced by Feldman \shortcite{feldman2010embodied}, Narayanan \shortcite{narayanan2010mind}, Bergen \shortcite{bergen2012louder}, and Evans \shortcite{evans2013language}) argues that  language comprehension is accomplished by means of such mind reading operations. Similarly, within psychology, there is an established body of work arguing for ``mental simulations" of future or possible outcomes, as well as interpretations of perceptual input \cite{graesser1994constructing,barsalou1999perceptions,zwaan1998situation,zwaan2012revisiting}.
These simulation approaches can be referred to as {\it embodied theories of mind}. 
 %
 Their goal is to view the semantic interpretation of an expression by means of a simulation, which is either mental (a la Bergen and  Evans) or  interpreted graphs such as Petri Nets (a la Narayanan and Feldman).  
 

In this position paper, we introduce a simulation framework, VoxWorld, that integrates the functionality and the goals of all three approaches above. Namely, we  situate an embodied agent in a multimodal simulation, with {\it mind-reading} interpretive capabilities, facilitated
through assignment and evaluation of object and context parameters within the environment being modeled.

\begin{figure}[htbp]
    \centering
    \includegraphics[width=0.18\textwidth]{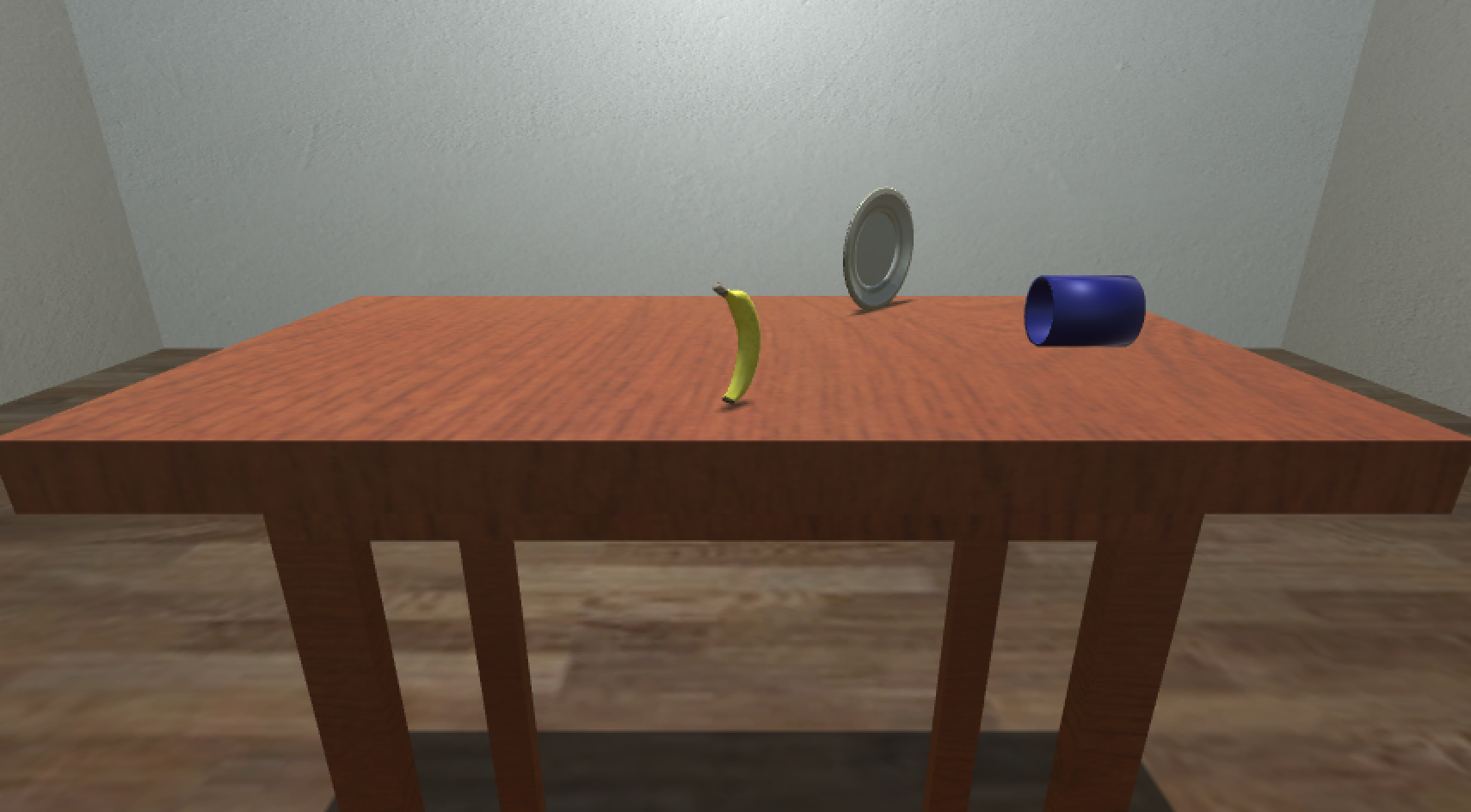}
    \includegraphics[width=0.18\textwidth]{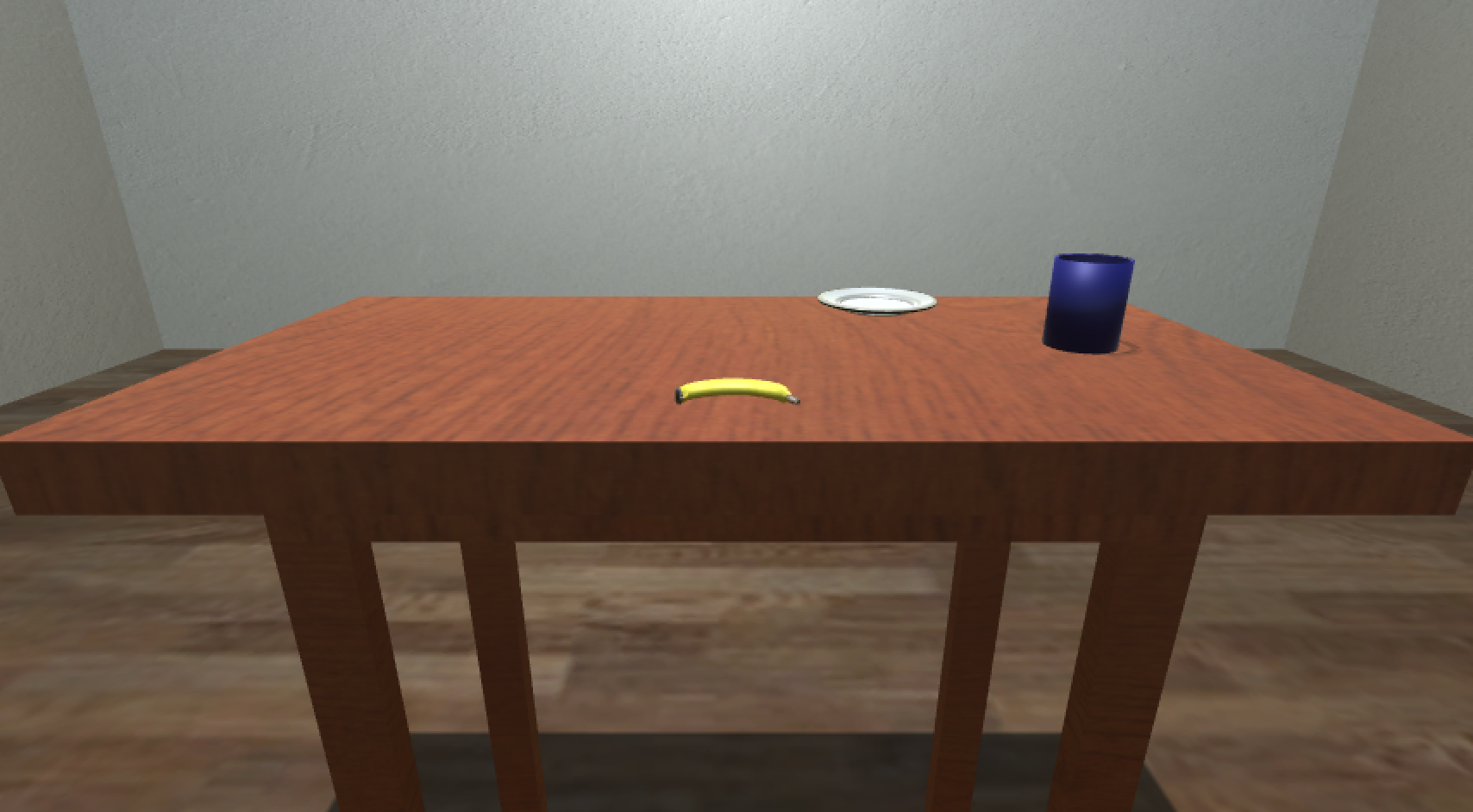}
    \caption{Objects in unnatural (L) and natural (R) positions}
    \label{fig:poses}
\end{figure}

For example, relations created by events persist after the completion of the event, and so in event simulation, they must also persist in order for the simulation to be considered accurate.  Fig.~\ref{fig:poses} shows a number of objects at similar locations but in one case in orientations that, due to the effects of physics, would be considered ``unstable" after the completion of a placement event.  Object knowledge about thinks like {\it shape of cup}, {\it top of plate}, {\it default position of banana} mean that human observers can judge the image on the left to be unsatisfactory results of placement events and that on the right to be more prototypical, due to the human ability to {\it simulate} what the result of an event in a given environment likely will be.

VoxWorld is based on the semantic scaffold provided by the VoxML modeling language \cite{PUSTEJOVSKY16.1101}, which provides a dynamic, interpretable model of objects, events, and their properties.  This allows us to create visualized simulations of events and scenarios that are rendered analogues to the ``mental simulations" discussed above.  VoxSim \cite{krishnaswamy2016multimodal,krishnaswamy2016voxsim} serves as the event simulator within which these simulations are created and rendered in real time, serving as the computer's method of visually presenting its interpretation of a situation or event.
Because modalities are modes of presentation, a multimodal simulation entails as many presentational modes as there are modalities being modeled. 
The visual modality of presentation (as in embodied gaming) necessitates ``situatedness'' of the agent,  as do the other perceptual modalities. Therefore, when we speak of {\it multimodal simulations}, they are inherently situated.
In a human-computer interaction using such a simulation, the simulation is a demonstration of the computational agent ``mind-reading" capabilities (an {\it agent simulation}). 
If the two are the same (where the agent is a proxy for the player or user, then the ``mind-reading" is just a demonstration of the scenario)
If, on the other hand, the two are separate (agent is {\it not} proxy for the user), then the simulation/demonstration communicates the agent's understanding of the user and the interaction. In this case, this demonstration entails the illustration of  both epistemic and perceptual content of the agent.

We believe that simulation can 
play a crucial role in  human-computer communication; it creates a shared epistemic model of the environment inhabited by a human  and an artificial agent, and demonstrates the knowledge held by the agent publicly. 
Demonstrating knowledge is needed to ensure a shared understanding  with its human interlocutor.
 If an agent is able to receive information from a human and interpret that relative to its current physical circumstances, it can create an epistemic representation of that same information.  However, without a modality to express that representation independently, the human is unable to verify or query what the agent is perceiving or how that perception is being interpreted.  In a simulation environment the human and computer share an epistemic space, and any modality of communication that can be expressed within that space (e.g., linguistic, visual, gestural) enriches the number of ways that a human and a computer can communicate within object and situation-based tasks, such as those investigated by  
Hsiao et al. \shortcite{hsiao2008object}, Dzifcak et al. \shortcite{dzifcak2009and}, and Cangelosi \shortcite{cangelosi2010grounding}, among others.
 
VoxWorld, and the accompanying simulation environment provided by VoxSim, includes the perceptual domain of objects, properties, and events. In addition, propositional
content in the model is accessible to the simulation.  Placing even a simple scenario, such as a blocks world setup, in a rendered 3D environment opens the search space to the all the variation allowed by an open world, as objects will almost never be perfectly aligned to each other or to a grid, with slight offsets in rotation caused by variations in interpolation, the frame rate, or effects of the platform's physics.  Nevertheless, when the rendering is presented to a user, the user can use their native visual faculty to quickly arrive at an interpretation of what is being depicted.

Situational embodiment takes place in real time, so in the case of a situation where there may be too many variables to predict the state of the world at time $t$ from a set of initial conditions at time $0$, situational embodiment within the simulation allows the reasoning agent to reason forward about a specific subset of consequences of actions that may be taken at time $t$, given the agent's current conditions and surroundings. 
Situatedness and embodiment is required to arrive at a complete, tractable interpretation given any element of non-determinism.  For example, an agent trying to navigate a maze from start to finish could easily do so with a map that provides them complete, or at least sufficient, information about the scenario.  If, however, the scene includes a disruptor (e.g., the floor crumbles, or doors open and shut randomly), the agent would be unable to plot a course to the goal.  It would have to start moving, assess the current circumstances at every timestep, and choose the next move or next set of $n$ moves based on them.  Situated embodiment allows the agent to assess next move based on the current set of relations between itself at the environment (e.g., ability to move forward but not leftward at the current state).  This provides for reasoning that not only saves computational resources but performs more analogously to human reasoning than non-situated, non-embodied methods.

\section{A Formal Interpretation of Simulations}

Given the distinction above between interpretations for ``simulation," we have been developing an approach that integrates all three: a situated embodied environment built on a game engine platform.  The computer, either as an embodied agent distinct from the viewer, or as the totality of the rendered environment itself, presents an interpretation ({\it mind-reading}) of its internal model, down to specific parameter values, which are often assigned for the purposes of testing that model.

We assume that a simulation is a contextualized 3D virtual realization of both the situational environment and the co-situated agents, as well as the most salient content denoted by communicative acts in discourse between them.  VoxWorld and VoxML \cite{PUSTEJOVSKY16.1101}, provide the following characteristics: object encoding with rich semantic typing and action affordances; action encoding as multimodal programs; it reveals the elements of the common ground in interaction between parties, be they humans or artificially intelligent agents. 

``Common ground" in a computational context relies on implementations of the following: 

\begin{enumerate}
    \item {\it Co-situatedness} and {\it co-perception} of the agents, such that they can interpret the same situation from their respective frames of reference.  This might be a human and an avatar perceiving the same virtual scene from different perspectives (see Fig.~\ref{fig:env-still}), or a combined virtual-physical scene with the integration of computer vision technology; or a human sharing the perspective of a robot as it navigates through a disaster zone.
    \item {\it Co-attention} of a shared situated reference, which allows more expressiveness in referring to the environment (i.e., through language, gesture, visual presentation, etc.).  The human and avatar might be able to refer to objects on the table in multiple modalities with a common model of differences in perspective-relative references (e.g., ``your left, my right"); or the human sharing the robot's perspective might be able to direct its motion using reference in natural language (``go through the second door on the left") or gesture (``go this way," with pointing).
    \item {\it Co-intent} of a common goal, such that adversarial relationships between agents reflect a breakdown in the common ground.  Here, the human and avatar in interaction around a table might seek to collaborate to build a structural pattern known to one or both of them; or the human and robot sharing perspective both have a goal to free someone trapped behind a door in a fire.  The robot informs the human about the situation and the human helps the robot problem-solve in real time until the goal is achieved.
\end{enumerate}

The theory of common ground has a rich and diverse literature concerning what is shared or presupposed in human communication \cite{clark1991grounding,gilbert1992social,stalnaker2002common,asher1998common,tomasello2007shared,PustejovskyCommonGround}.

VoxML (Visual Object Concept Markup Language) forms the scaffold used to encode knowledge about objects, events, attributes, and functions by linking lexemes to their visual instantiations, termed the ``visual object concept" or {\it voxeme}.  In parallel to a lexicon, a collection of voxemes is termed a {\it voxicon}.  There is no requirement on a voxicon to have a one-to-one correspondence between its voxemes and the lexemes in the associated lexicon, which often results in a many-to-many correspondence.  That is, the lexeme {\it plate} may be visualized as a {\sc [[square plate]]}, a {\sc [[round plate]]}, or other voxemes, and those voxemes in turn may be linked to other lexemes such as {\it dish} or {\it saucer}.

Each voxeme is linked to either an object geometry, a program in a dynamic semantics, an attribute set, or a transformation algorithm, which are all structures easily exploitable in a rendered simulation platform. 

An {\sc object} voxeme's semantic structure provides {\it habitats}, which are situational contexts or environments  conditioning the object's {\it affordances}, which may be either ``Gibsonian" affordances \cite{gibson1982reasons} or ``Telic'' affordances \cite{Pustejovsky1995,pustejovsky2013dynamic}. A habitat specifies how an object typically occupies a space. When we are challenged with computing the embedding space for an event, the individual habitats associated with each participant in the event will both define and delineate the space required for the event to transpire.  Affordances are used as attached behaviors, which the object either facilitates by its geometry (Gibsonian) or purposes for which it is intended to be used (Telic).  For example, a Gibsonian affordance for [[{\sc cup}]] is ``grasp," while a Telic affordance is ``drink from."   This allows procedural reasoning to be associated with habitats and affordances, executed in real time in the simulation, inferring the complete set of spatial relations between objects at each frame and tracking changes in the shared context between human and computer.  

It also allows the system to reason about objects and actions independently.  When simulating the objects alone, the simulation presents how the objects change in the world.  By removing the objects and presenting only the actions that the viewer would interpret as {\it causing} the intended object motion (i.e., a pantomime of an embodied agent moving an object without the object itself), the system can present a ``decoupled" interpretation of the action, for example, as an animated gesture that traces the intended path of motion.  By composing the two, it demonstrates that particular instantiation of the complete event.  This allows an embodied situated simulation approach to easily compose objects with actions by directly interpreting at runtime how the two interact.

\section{Reasoning within an Interpreted Simulation}

\begin{figure}[htbp]
    \centering
    \includegraphics[width=0.1\textwidth]{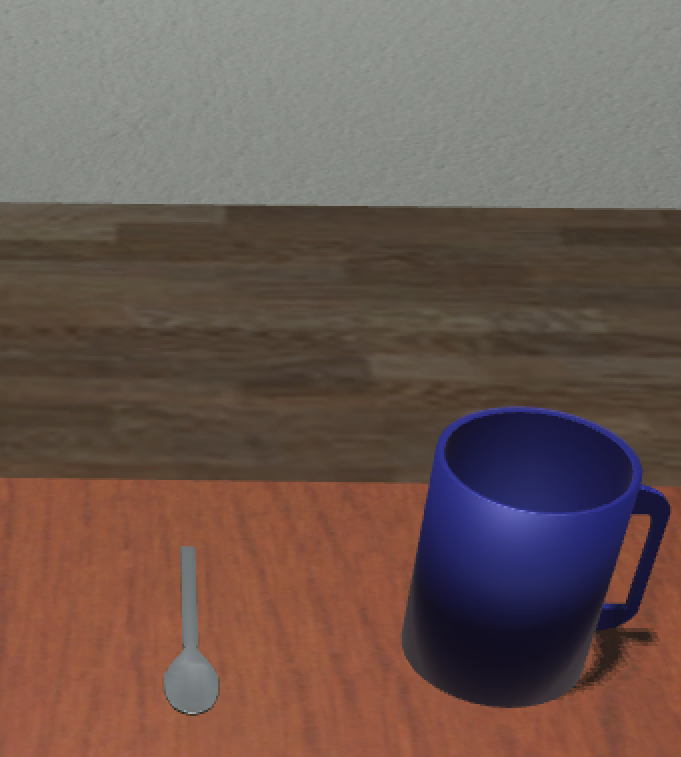}
    \includegraphics[width=0.1\textwidth]{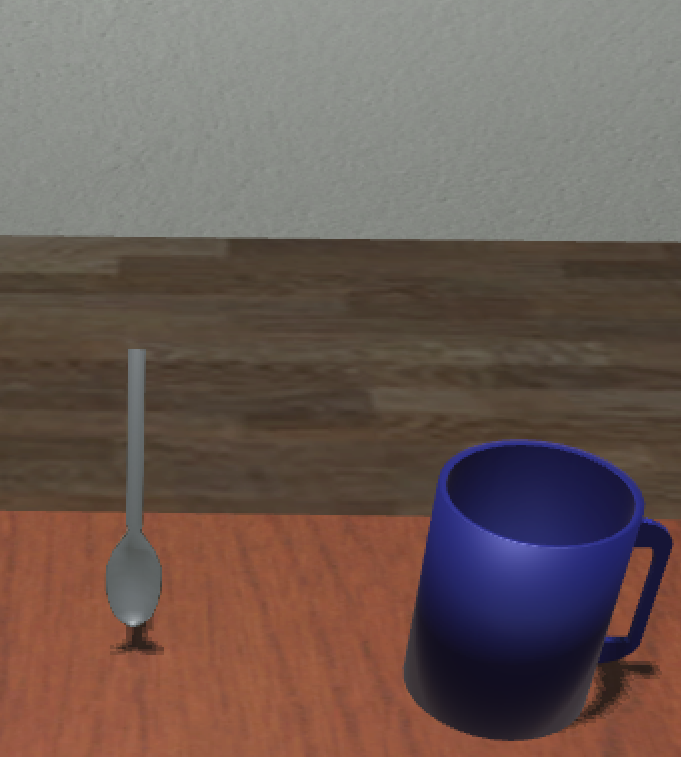}
    \includegraphics[width=0.1\textwidth]{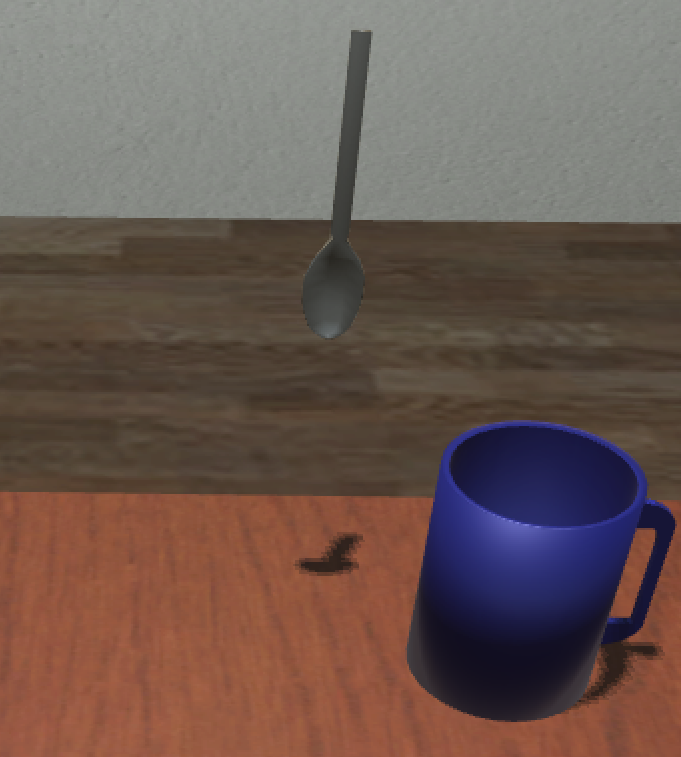}
    \includegraphics[width=0.1\textwidth]{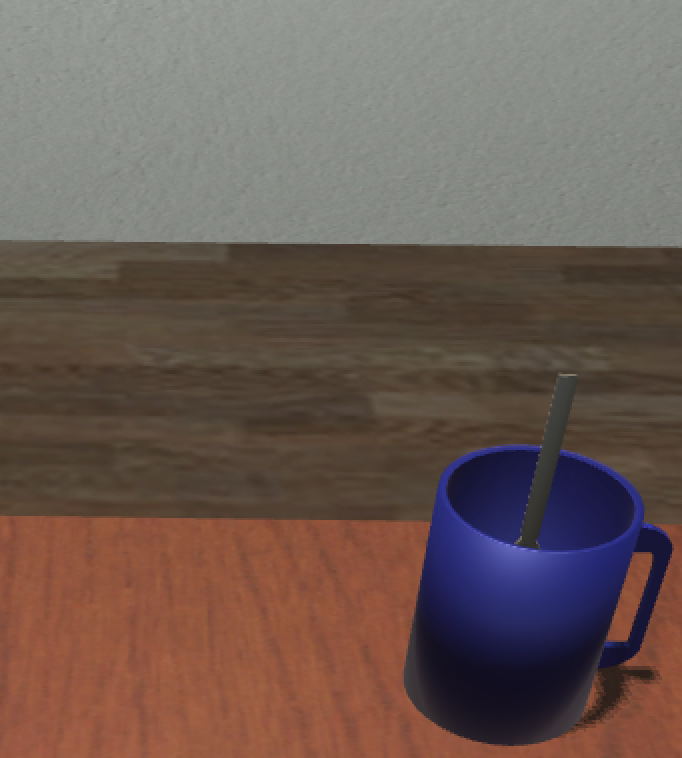}
    \caption{[[{\sc spoon in mug}]]}
    \label{fig:knife_in_mug}
\end{figure}

{\bf VoxSim} \cite{krishnaswamy2016multimodal,krishnaswamy2016voxsim} implements the VoxML platform in the Unity game engine software by Unity Technologies\footnote{https://unity3d.com/}.  The current implementation of VoxSim provides scenes in a Blocks World domain, augmented with a set of more complicated or interesting everyday objects (e.g., cups, plates, books, etc.).  There are scenes without an avatar where the user can direct the computer to manipulate objects in space (see Figs.~\ref{fig:poses} and ~\ref{fig:knife_in_mug}) or with an avatar that can act upon objects and respond to the user's input where it is ambiguous (see Fig.~\ref{fig:env-still}).  {\it VoxWorld} contains other software, models, and interfaces, e.g., to consume input from CNN-based gesture recognizers \cite{krishnaswamy2017communicating}, and to track and update the agent's epistemic state or knowledge about what the human interlocutor knows.

It is a straightforward process to create new scenes with 3D geometries with packaged code that handles the creates and instantiation of voxemes, handles their interactions and performs basic spatial reasoning over them.  We also provide a library of basic motion predicates and methods of composing them into more complex actions using VoxML.

Given the continuous tracking of object parameters such as position and orientation, facilitated by a game engine or simulation, and the knowledge of object, event, and functional semantics facilitated by a formal model, an entity's interpretation at runtime can be computed in conjunction with the other entities it is currently interacting with and their properties.  One such canonical example would be placing an object [[{\sc spoon}]] in an [[{\sc in}]] relation with another object [[{\sc mug}]] (Fig.~\ref{fig:knife_in_mug}).

The mug has an intrinsic top, which is aligned with the upward Y-axis of the world or embedding space (denoted in VoxML as \{$align(Y,\mathcal{E}_{Y})$, $top(+Y)$\}).  The mug is also a concave object, and the mug's geometry (the [[{\sc cup}]], excluding the handle) has reflectional symmetry across its inherent (object-relative) XY- and YZ-planes, and rotational symmetry around its inherent Y-axis such that when the object is situated in its inherent $top$ habitat, its Y-axis is parallel to the world's.  From this we can infer that the {\it opening} (e.g., access to the concavity) must be along the Y-axis.  Encoding the object's concavity also allows fast computation for physics and collisions using bounding boxes, while still facilitating reasoning over concave objects.

VoxSim performs reasoning over 3D variants of well-known spatial calculi (e.g., Albath et al. \citeyear{albath2010rcc}) and  interval/point calculi from libraries such as QSRLib \cite{gatsoulis2016qsrlib}, computing axial overlap with the Separating Hyperplane Theorem \cite{schneider2014convex}.  In order to put object $x$ $in$ object $y$, while maintaining external contact with $y$'s concave geometry, the placed object $x$ must {\it fit inside} the concave object $y$.  In the case of the mug, it can be reasoned as shown that its concavity opens along the Y-axis, so any computational reasoner must also determine that the object to be placed within it can fit in that same orientation.  In the case of a spoon, normally lying flat on a surface, somewhere flush with the world's XZ-plane, simply placing it at the point where it would touch the bottom of the inside of the mug would also cause it to interpenetrate the mug's sides inappropriately, and so the spoon must first be turned (rotated) to align with the mug's opening.  The requirements on simulating {\it put the spoon in the mug} enforce the resulting state of this ``turn" action as a precondition, which allows for intelligent decision making typically not learnable from a modality such as language or still images.

Deformation of the object is also possible, as long as it maintains the object's topological isomorphism.  Just as individual transformations over rigid object bounding boxes can be tested for relation or event satisfaction in a given context, transformations over individual vertices and edges can be performed to search for the set of deformations that satisfy a known constraint.  For instance, the bounds of an object $x$ and a containing space $y$ can be extracted at each frame as deformations are performed over $x$ such that a predicate like as $contains(extents(x),extents(y))$ can be computed to test the satisfaction condition of $put(x,in(y))$ where $x$ is deformable.  This allows us, within VoxSim, to solve for deformations that describe events such as {\it crumple} or {\it fold}.  Even within the continuous open world of a game or simulation-based environment, searching for transformations over individuals within a finite set of vertices and edges that satisfy a predetermined condition keeps the search space tractable well within computational limits and facilitates the gathering of data for experiments that can teach an AI agent to solve decidable problems such as ``how to fold a cloth" or ``how to fit numerous items in a container."


Search can be performed in an embodied, situated simulation environment through parameter setting of the type found in {\it computational simulation modeling}.  In order to generate a visualization of an event, all variable parameters must have values assigned, otherwise the program will fail to run.  This requirement on the game engine software also becomes a requirement on the creation of a fully-defined simulation model.  The composition of objects and events provide much of the needed information but in cases where parameters still require values (e.g., the speed of a moving object described simply by the predicate ``slide"), we can use Monte Carlo value assignment to set those values in the simulation environment.  The rendered simulation including the stochastically-assigned value(s) is presented as the system's interpretation of that situation being modeled formally.

We have performed a number of experiments in this area using the VoxSim software \cite{krishnaswamy2017montecarlo}.  For a set of motion predicates where various parameters are left underspecified in the linguistic description (e.g., ``slide the block across the table" says nothing about speed or direction of motion; ``put the block next to book" or ``touch the block to the book" does not fully specify the relative placement of the two objects in 3D space), we generated multiple simulations of such events, captured them on video along with the specific parameter values used in each simulation, and had human evaluators choose the best simulation out of three for one description, and the best description out of three for one simulation.  This allowed us generate a novel dataset of motion events and parameter values descriping prototypical instances of them (according to evaluators), and such data can be used to train a model that captures contextual dependecies for better simulation generation and interpretation.

\section{Learning by Communication}

One of the things that an embodied simulation model for AI enables is {\it peer-to-peer communication}, specifically because of the requirement that the AI agent have some kind of situated embodiment in which it interprets its environment.  This allows the creation of common ground between the human and the AI that allows them to communicate \cite{pustejovsky2017creating}.

\begin{figure}
\centering
\includegraphics[width=2in]{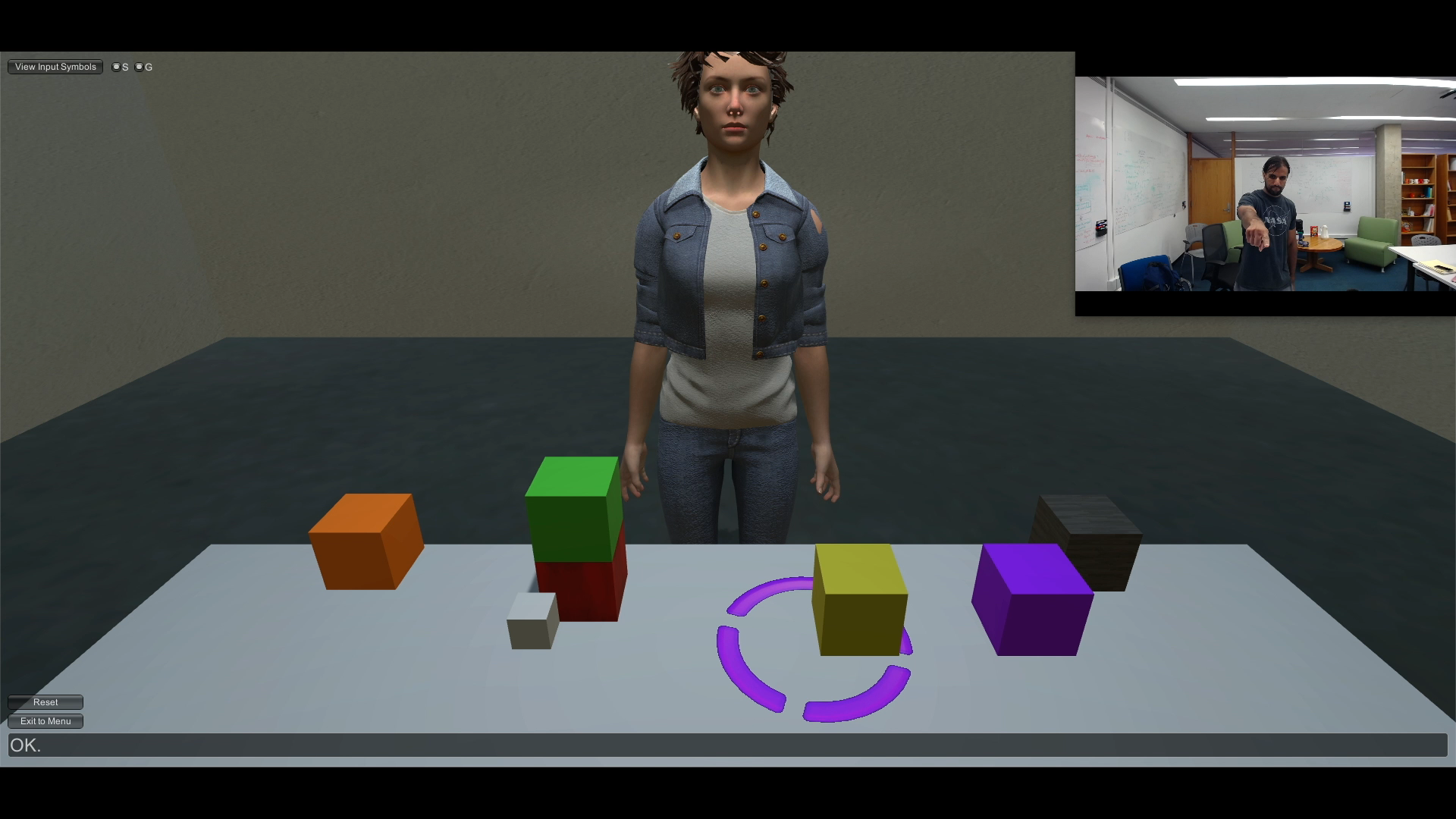}
\caption{\label{fig:env-still}Screenshot of human-avatar interaction in VoxSim. The purple circle indicates the location interpreted as the target of the human subject's pointing.}
\end{figure}

We use the example of learning to build a structure, namely a staircase. In a series of user studies, we had naive human subjects interact with an avatar (a screenshot of VoxSim is given in Fig.~\ref{fig:env-still}) to build a staircase out of six blocks.  Due to subjects' lack of skill using the system fluently, the generated structures all satisfied the user's notion of a staircase, but across the 17 samples were often diverse and noisy (variants included spaces between blocks, blocks not properly aligned, or blocks rotated).  Each of the 17 structures is defined by $\approx$20 qualitative spatial relations, and each set is stored as an unordered list.

\begin{figure}
\centering
\includegraphics[height=.6in]{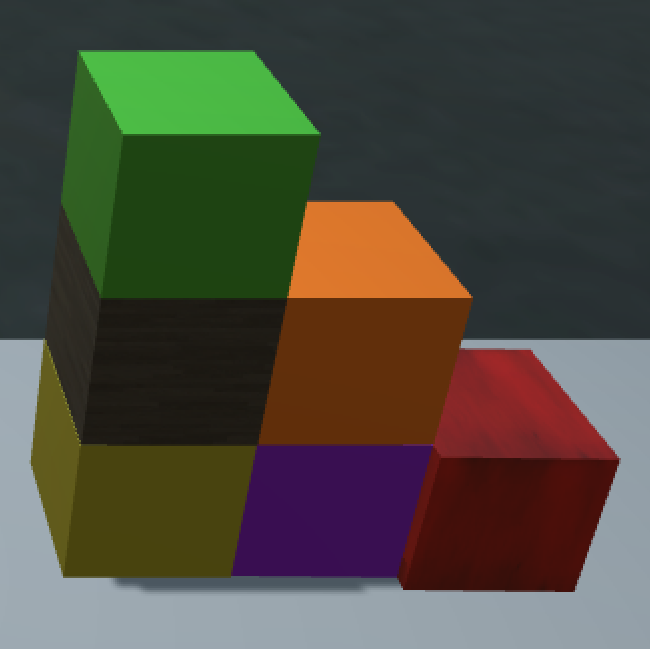}
\includegraphics[height=.6in]{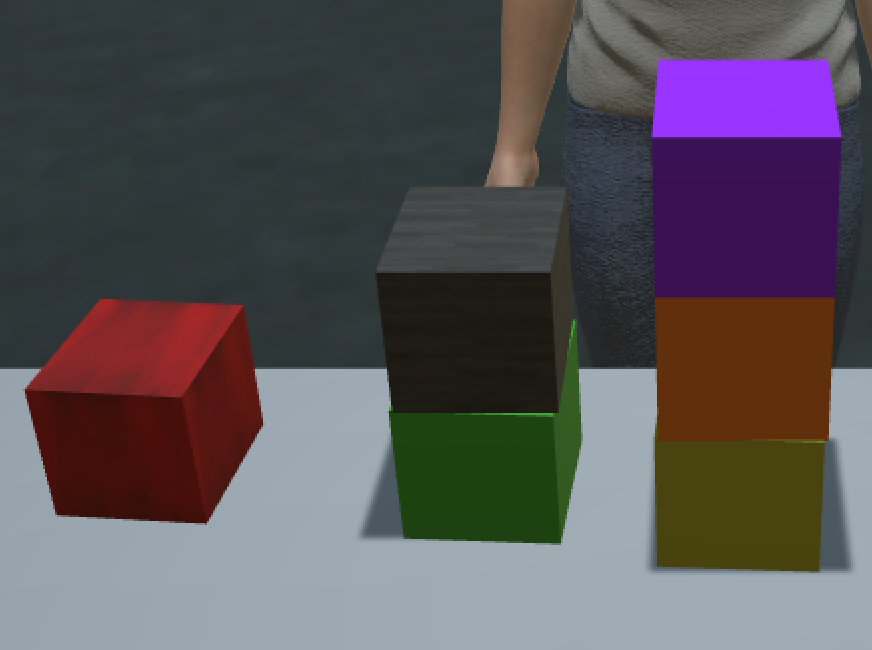}
\includegraphics[height=.6in]{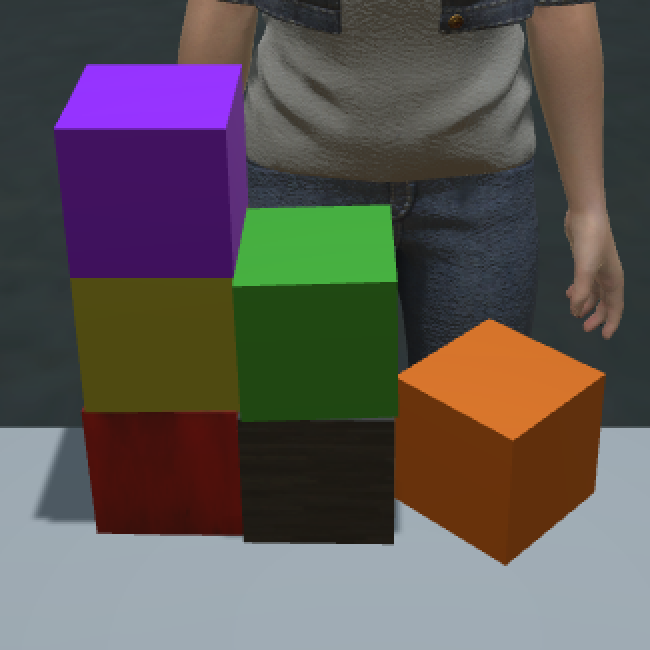}
\caption{\label{fig:examples}Example user-constructed staircases}
\end{figure}

The qualitative relation sets that defined each structure, extracted directly from VoxSim via Unity, were then used to train a model for the avatar to use to build its own novel instance of a staircase structure.  The avatar in the simulation places a block then, using a 4-layer convolutional neural network, chooses one of its known examples to begin approximating with the next step.  Using a long short-term memory network of 3 layers with 32 nodes each, and trained up to 20 timesteps, the agent selects the most likely sequence of moves that would approximate the chosen example structures.  These moves are then pruned using heuristic graph-matching and a move is chosen. Further details of the method and sample results are given in \cite{krishnaswamy2019combining}.

\begin{figure}
\centering
\includegraphics[height=.6in]{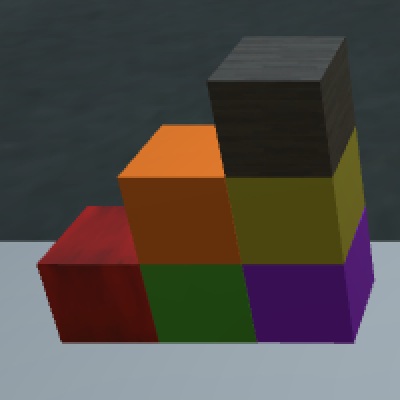}
\includegraphics[height=.6in]{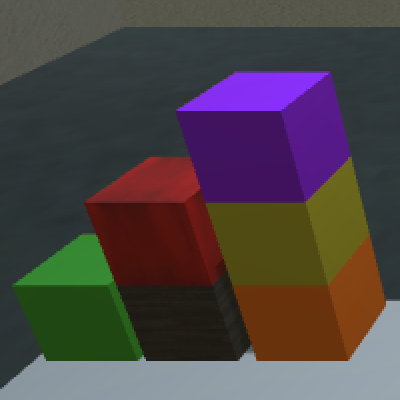}
\includegraphics[height=.6in]{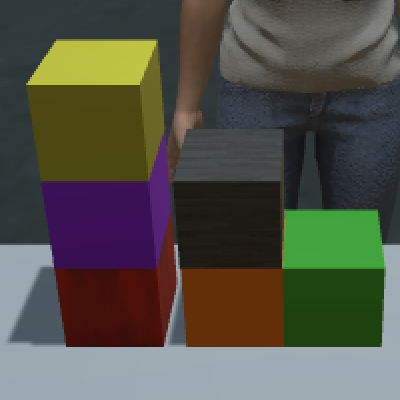}
\includegraphics[height=.6in]{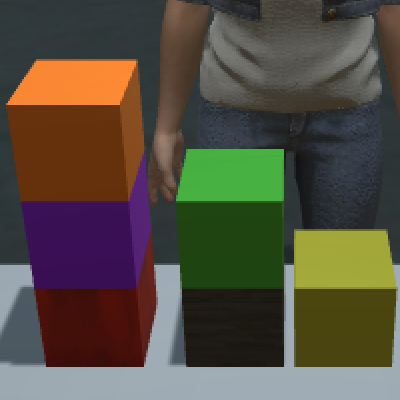}
\caption{\label{fig:examples}Example learned staircases}
\end{figure}

However, sometimes the agent generates a structure that does not comport with an observer's understanding of that structure label.  In previous experimentations using a similar framework for action learning \cite{do2018teaching} we encountered difficulty in using negative examples to steer the learning agent away from incorrect generations (cf. also \cite{dietterich2000ensemble,nguyen2015deep}).  Here, the interaction within the simulation environment facilitates ``learning by communication" that allows us to take a negative example and turn it into a corresponding positive example, and  storing both increases the overall data size and gives a clear minimal pair between a good example and a bad one.

\begin{figure}
\centering
\includegraphics[width=1in]{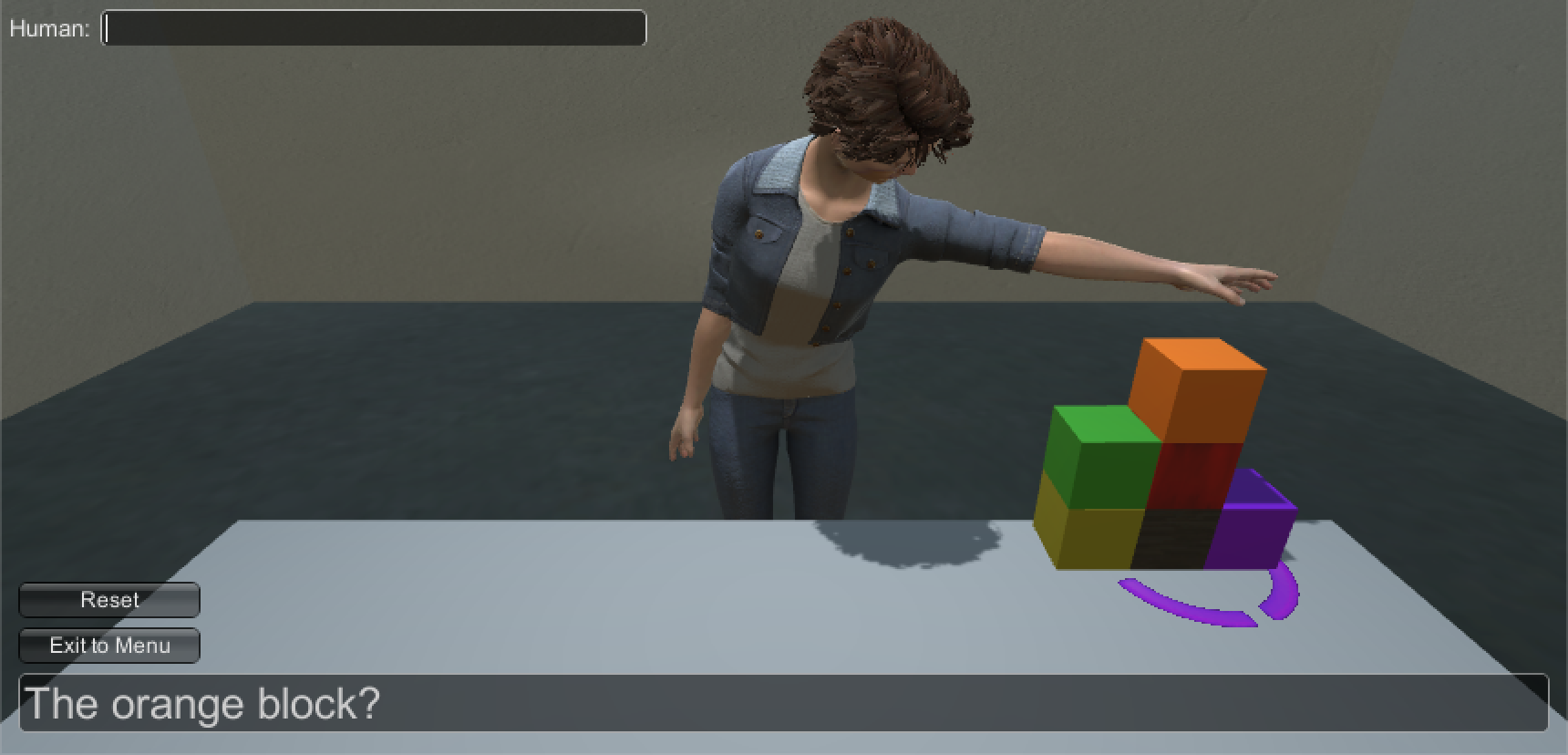}
\includegraphics[width=1in]{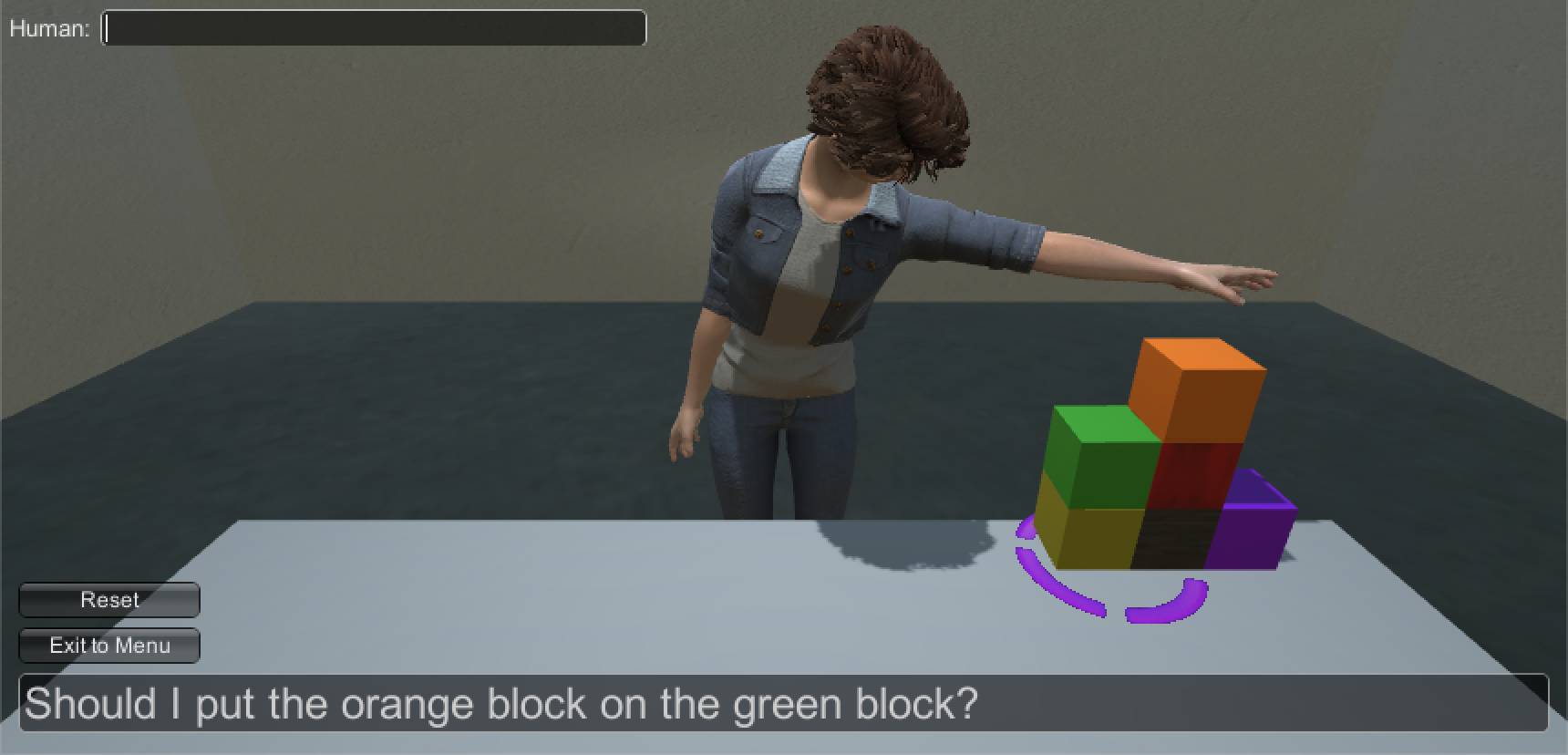}\\
\includegraphics[width=1in]{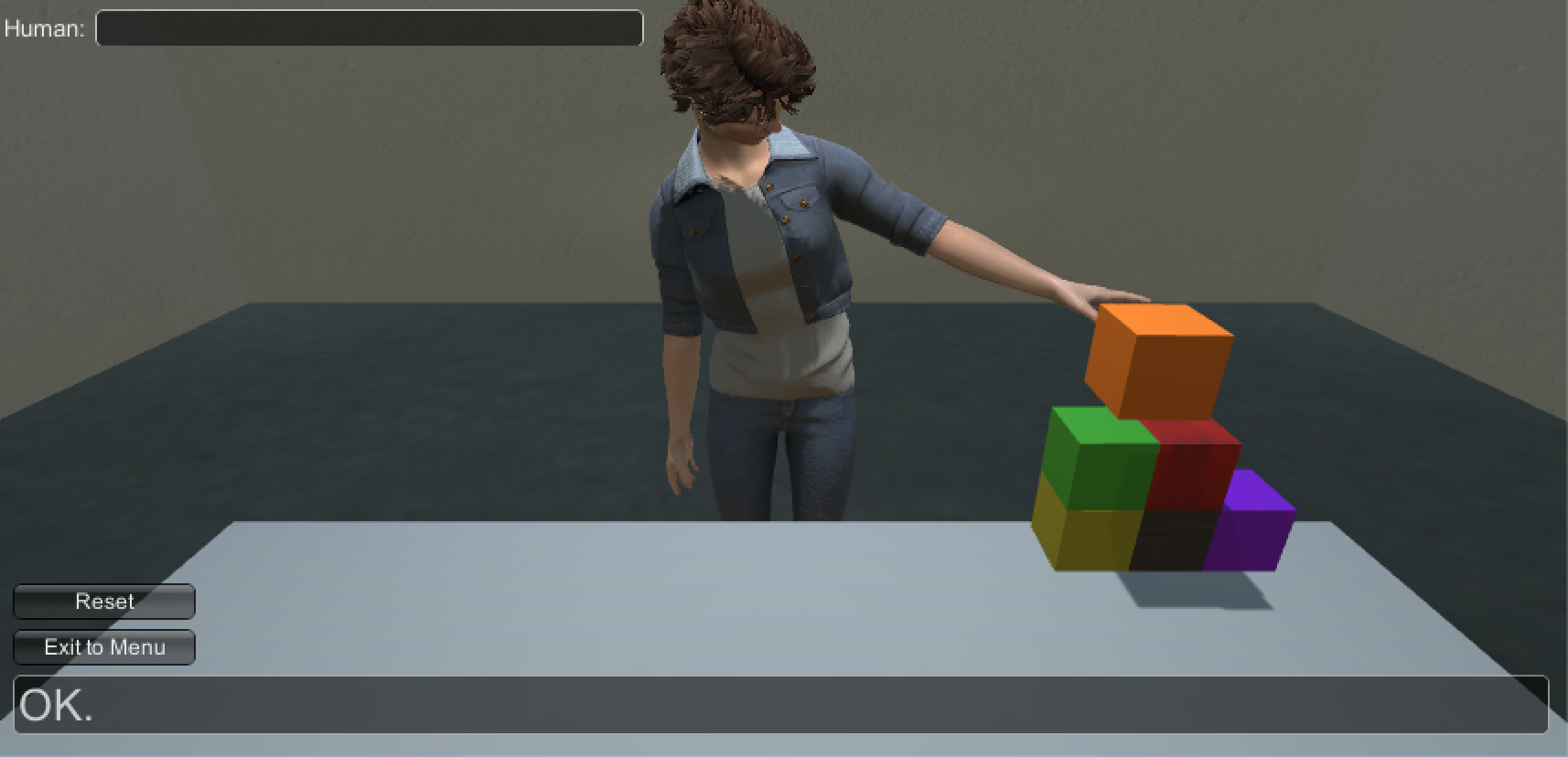}
\includegraphics[width=1in]{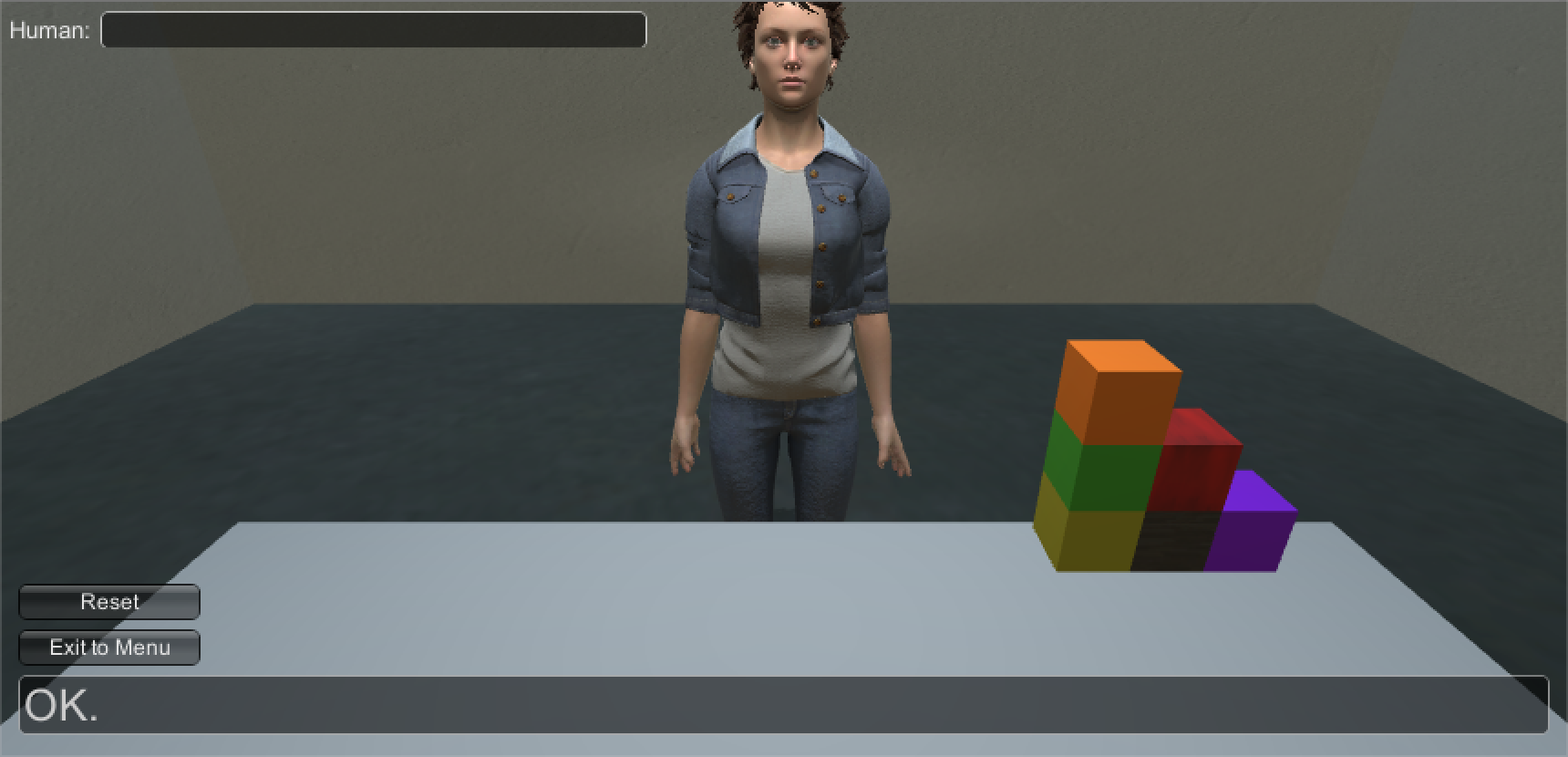}
\caption{\label{fig:correction}Correcting a generated structure with multimodal communication.}
\end{figure}

A correction interaction might proceed as follows:
\begin{enumerate}
\item The system generates the ``staircase" in the top left image shown in Fig.~\ref{fig:correction}, which is one block off from a prototypical staircase.  This gets marked as an incorrect case.
\item The user points to the orange block at the top of the structure.  The agent clarifies this request.
\item The user indicates the green block and gestures for (or tells) the agent to move the orange block there.
\item The agent clarifies this request and makes the move.  The result is marked as a correct case.
\end{enumerate}

This is demonstrated in the simulation environment. Because of the simulation acting in the service of multimodal communication between a human and a computer, we can improve the model (understanding) of the computer's concepts (here, a staircase) through learning by communication, and iterative demonstration of the agent's model.

\section{Conclusion}

Across different fields and in the existing AI, cognition, and game development literature, there exist many different definitions of ``simulation."  Nonetheless, we believe the common thread between them is that simulations as a framework facilitate both qualitative and quantitative reasoning by providing quantitative data (for example, exact coordinates or rotations) that can be easily converted into qualitative representations.  This makes simulation an effective platform for both producing and learning from datasets.

When combined with formal encodings of object and event semantics, at a level higher than treating objects as collections of geometries, or events as sequences of motions or object relations, 3D environments provide a powerful platform for exploring ``computational embodied cognition."  Recent developments in the AI field have shown that common-sense understanding in a general domain requires either orders of magnitude more training data than traditional deep learning models, or more easily decidable representations, involving context, differences in perspective, and grounded concepts, to name a few.

Technologies in use in the gaming industry are proving to be effective platforms on which to develop systems that afford gathering both traditional data for deep learning and representations of common sense, situated, or embodied understanding.  In addition, game engines perform a lot of ``heavy lifting," providing APIs for UI and physics, among others, which allows researchers to focus on implementing truly novel functionality and develop tools for experimentation in simulation-based and qualitative understanding of both human and machine cognition and intelligence.

\bibliographystyle{aaai}
\bibliography{References}

\end{document}